\documentclass[runningheads]{llncs}

\usepackage{eccv}

\usepackage{eccvabbrv}

\usepackage{graphicx}
\usepackage{booktabs}

\usepackage[accsupp]{axessibility}  

\usepackage[pagebackref,breaklinks,colorlinks,citecolor=eccvblue]{hyperref}

\usepackage{orcidlink}

\usepackage{xcolor}
\usepackage{multirow}
\usepackage{subcaption}
\usepackage{amsmath}
\usepackage{arydshln}
\usepackage{wrapfig}
\usepackage{float}

\DeclareMathOperator*{\argmax}{arg\,max}

\newcommand{\methodname}{PALM\xspace}

\newcommand{\myparagraph}[1]{\noindent\textbf{#1.}}

\begin{document}

\title{PALM: Predicting Actions through Language Models} 


\author{Sanghwan Kim\inst{1} \and 
Daoji Huang\inst{1} \and 
Yongqin Xian\inst{2} \and \\
Otmar Hilliges\inst{1} \and 
Luc Van Gool\inst{1,3,4} \and
Xi Wang\inst{1}}

\authorrunning{S. Kim et al.}

\institute{ETH Zürich \and Google \and KU Leuven \and INSAIT, Sofia}

\maketitle

\begin{abstract}
Understanding human activity is a crucial yet intricate task in egocentric vision, a field that focuses on capturing visual perspectives from the camera wearer's viewpoint. 
Traditional methods heavily rely on representation learning that is trained on a large amount of video data. 
However, a major challenge arises from the difficulty of obtaining effective video representation. 
This difficulty stems from the complex and variable nature of human activities, which contrasts with the limited availability of data. 
In this study, we introduce \methodname, an approach that tackles the task of long-term action anticipation, which aims to forecast forthcoming sequences of actions over an extended period.
%
%
Our method \methodname incorporates an action recognition model to track previous action sequences and a vision-language model to articulate relevant environmental details. By leveraging the context provided by these past events, we devise a prompting strategy for action anticipation using large language models (LLMs). 
Moreover, we implement maximal marginal relevance for example selection to facilitate in-context learning of the LLMs. 
Our experimental results demonstrate that \methodname surpasses the state-of-the-art methods in the task of long-term action anticipation on the Ego4D benchmark. 
We further validate \methodname on two additional benchmarks, affirming its capacity for generalization across intricate activities with different sets of taxonomies. 


\end{abstract}    
\section{Introduction}
\label{sec:introduction}

As wearable technology, such as smart glasses and body-worn cameras, becomes more prevalent, the availability of egocentric videos has grown significantly. These videos capture activities from the camera wearer’s viewpoint, providing a unique perspective that features individual behaviors in various contexts. Egocentric vision holds immense promise for understanding how humans execute specific tasks and activities. It enables numerous applications such as real-time systems that anticipate and respond to user needs~\cite{rodin2021predicting, yao2019egocentric, ryoo2015robot}, monitoring patient activities~\cite{zhan2014multi, nakazawa2019first, meditskos2018multi}, and providing personalized instructions and procedures~\cite{ohnbar2018personalized, neumann2019future, leo2017computer}.

\begin{figure*}[t]
\centerline{\includegraphics[scale=.35]{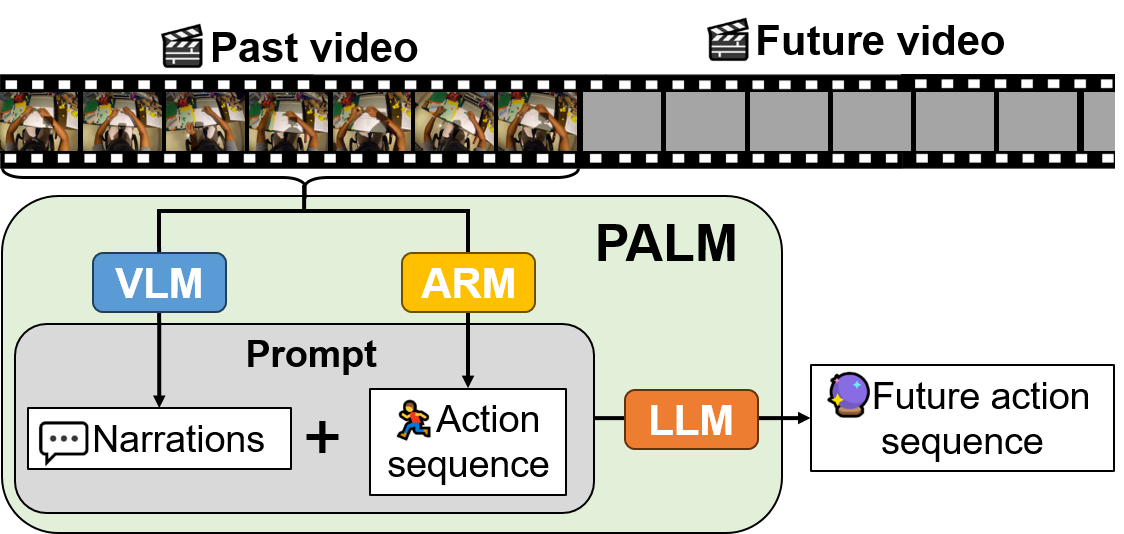}}
\caption{\textbf{\methodname for long-term action anticipation}. Given a video input, \methodname uses a vision-language model (VLM) and an action recognition model (ARM) to compile a text prompt that describes the past video content. A large-language model (LLM) then takes the prompt and predicts future actions.}
\label{fig:teaser_figure}
\vspace{-0.5cm}
\end{figure*}

Existing works on understanding human activities mostly rely on learning representations from large-scale video datasets~\cite{pramanick2023egovlpv2, wang2022internvideo, ragusa2023stillfast, fan2021multiscale}. However, these approaches face two critical challenges. First, the complexity and variability of human activities make it challenging to obtain comprehensive representations, particularly for longer videos. Second, reliance on large datasets limits the model's capability to generalize to less common, long-tail classes and unseen scenarios. In this context, there is a pressing need for a more versatile approach to human activity understanding, adept at handling the complicated nature of human activities without overly relying on large-scale video data. Therefore, recent works have started to exploit the procedure knowledge and generalization ability of large language models (LLMs)~\cite{li2022pre, huang2022language, ahn2022can, patel2023pretrained, pasca2024summarize}.

Most prior works based on LLMs have been largely confined to generating instructions and plans given a high-level task description (e.g., "prepare the dinner table") and have not been tailored for intricate human activity understanding from videos. Indeed, there is a significant gap in the literature when it comes to the use of LLMs for grounded activity understanding, especially in challenges involving understanding long-term action sequences. In this work, we efficiently boost the versatile ability of LLMs in long-term action anticipation~\cite{grauman2022ego4d, damen2020epic, huang2023palm} by providing sufficient visual context in discrete text.

As shown in \cref{fig:teaser_figure}, we introduce a simple yet effective framework called \methodname that combines descriptions of past action sequences and corresponding narrations to bridge the gap between visual and textual domains. In detail, \methodname takes an input video with annotated action periods and leverages a vision-language model and an action recognition model to describe past events. We then design a prompt strategy for large language models (LLMs) to anticipate future actions given the context of these past events. To facilitate the predictive capabilities of LLMs, we use maximal marginal relevance (MMR)~\cite{ye2022complementary, carbonell1998use} to retrieve similar and diverse examples from the training set, enabling in-context learning for the LLMs. Through this prompting strategy, our system can tackle intricate tasks in understanding human activities, benefiting from existing foundation models. Experimental results demonstrate that our method \methodname significantly improves performance in long-term action anticipation, as evaluated using the Ego4D benchmark~\cite{grauman2022ego4d}. Further evaluations on two additional benchmarks, EPIC-KITCHEN~\cite{damen2020epic} and EGTEA~\cite{li2018eye}, reveal the generalization ability of \methodname across different contexts and datasets. Our contributions can be summarized as follows:

\vspace{-0.5em}
\begin{itemize}
    \item We propose an effective textural representation of past events, consisting of both action sequences and accompanying narrations that provide additional context and environmental details.
    \item We design a novel prompting strategy that enables efficient in-context learning for large language models.
    \item \methodname outperforms existing state-of-the-art methods in the task of long-term action anticipation, which is part of the Ego4D benchmark.
\end{itemize}

\section{Related Work}\label{sec:related_work}

\subsection{Foundation Model}

Foundation models serve as the cornerstone for various downstream tasks across different data modalities~\cite{zhou2023comprehensive, bommasani2021opportunities}. Here we consider two types of foundation models: vision-language models (VLMs) to generate the textual narrations of video and large language models (LLMs) to predict future action given visual context.

\myparagraph{Vision-language model}
VLMs utilize complementary characteristics of image and text datasets to perform multimodal tasks such as image captioning, visual question-answering, and visual reasoning~\cite{li2019visualbert, lu2019vilbert, radford2021learning}. Various works~\cite{zhou2020unified, cho2021unifying, wang2021simvlm} attempt to unify vision-language tasks with a single model, overcoming the limited utility of previous models. In this paper, the image or video captioning ability of VLMs is utilized to directly transform visual information into narrative text descriptions. We empirically found that BLIP family~\cite{li2022blip, li2023blip, instructblip} provides concise and informative captions. For instance, BLIP~\cite{li2022blip} proposes a novel encoder-decoder architecture specialized for multimodal pertaining, and BLIP-2~\cite{li2023blip} significantly reduces the trainable parameters by using a querying transformer to bridge the modality gap between frozen vision models and language models. InstructBLIP~\cite{instructblip} further performs instruction tuning on BLIP-2, enabling general-purpose models with a unified natural language interface.  

\myparagraph{Large language model}
Language foundation models~\cite{radford2019language, brown2020language, zhang2022opt, black2021gpt, touvron2023llama, touvron2023llama2} have had a significant impact on the NLP field, demonstrating surprising ability to generalize across unseen tasks. With their extensive training data and large parameter size, LLMs have demonstrated the ability to learn from examples provided in input prompts, a concept known as in-context learning~\cite{brown2020language}. LLMs have also shown success in the tasks of decision-making and planning. For example, SayCan~\cite{ichter2022can} combines a language model with learned low-level skills for grounded planning, while Socratic Models~\cite{zeng2022socratic} formulate action prediction as text completion problems from captioned images. 
In this work, we explore how such procedure knowledge can be exploited in the complex action anticipation task and showcase the effectiveness of in-context learning for this purpose.

\subsection{Long-term Action Anticipation}
Given an input video, the long-term action anticipation (LTA) task aims to predict the next actions in the future.
In this work, we explore various action recognition models to estimate accurate action sequences and extensively compare our method with previous action anticipation models.

\myparagraph{Action recognition}
SlowFast~\cite{feichtenhofer2019slowfast} proposes to fuse two channels of input that consist of a low frame and a high frame rate video, using 3D convolutional kernels. Similar to Slowfast, StillFast~\cite{ragusa2023stillfast} adopts a specific architecture with two branches to effectively process high-resolution still images and low-resolution video frames. On the other hand, MViT~\cite{fan2021multiscale} employs several channel-resolution scale stages to obtain a multiscale pyramid of features, based on vision transformers. Compared to previous works solely optimized on video inputs, EgoVLP~\cite{lin2022egocentric} and EgoVLPv2~\cite{pramanick2023egovlpv2} pre-train video and text encoders with contrastive loss specialized for the egocentric domain. Each encoder provides useful feature representations for downstream video tasks such as natural language queries, video question-answering, and video summarization.

\myparagraph{Action anticipation} 
Baselines provided by Ego4D~\cite{grauman2022ego4d} employ the SlowFast~\cite{feichtenhofer2019slowfast} video encoder to extract features from sampled clips, followed by transformer layers and classification heads to predict the verb-noun pair of future action sequence. HierVL~\cite{ashutosh2023hiervl} utilizes a hierarchical contrastive training objective that encourages text-visual alignment at both the clip and video levels. ICVAE~\cite{mascaro2023intention} proposes using a predicted scenario, which summarizes the activity depicted in the video, as a condition for action prediction. A few concurrent approaches~\cite{chen2023videollm, zhao2023antgpt, wang2023vamos} leverage the generalization abilities and abundant knowledge of LLMs to predict future actions. Compared to these LLM-based methods, \methodname differentiates itself by utilizing vision-language model for effective extraction of visual context and employing in-context learning in combination with MMR.

\section{Method}\label{sec:method}

\begin{figure*}[h]
\centerline{\includegraphics[scale=.45]{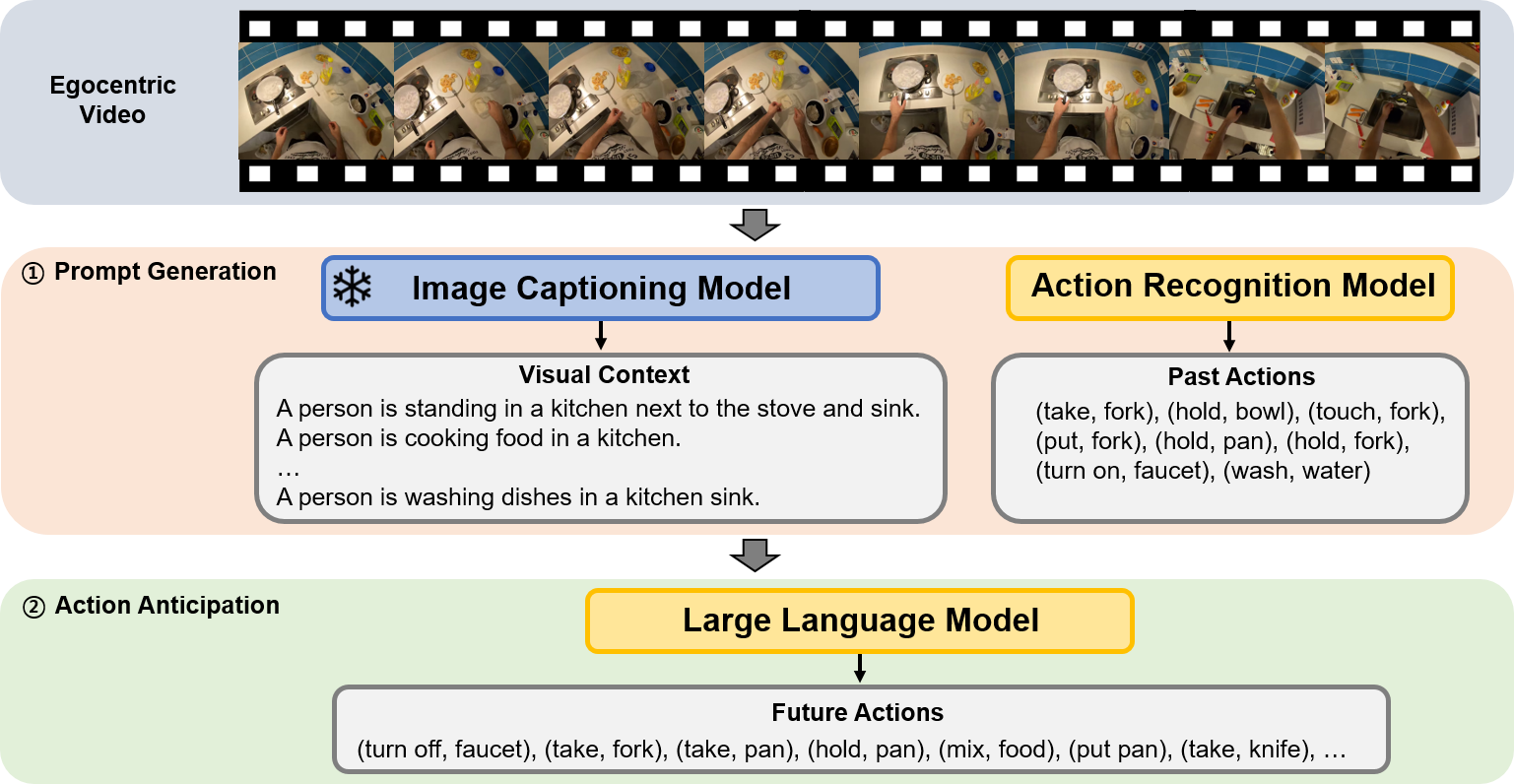}}
\caption{\textbf{Overview of \methodname}. Given an input video, an image captioning model and an action recognition model convert visual information of past activities into text format (visual context and past actions). Then, LLM utilizes visual context and past actions to predict the sequence of future actions. The image captioning model is frozen and in-context learning is adapted for LLM prompting.}
\label{fig:palm}
\vspace{-0.5cm}
\end{figure*}

We introduce \methodname, a comprehensive framework designed for video understanding and action anticipation. \methodname comprises three key modules, as illustrated in \cref{fig:palm}: an image captioning module (\cref{subsec:image_captioning}), an action recognition module (\cref{subsec:action_recognition}), and an action anticipation module based on LLMs (\cref{subsec:large_language}).

Given an untrimmed input video, the action recognition module generates labels for \emph{past actions} in the form of verb and noun pairs (eight past actions in \cref{fig:palm}). The action recognition model is trained to produce valid verb and noun pairs as predefined by each dataset. Simultaneously, the image captioning module generates captions from the input video, providing additional visual information relevant to the events, termed \emph{visual context}. The prompt is then constructed using the visual context and past actions, passed to LLMs to predict future actions. Additionally, we incorporate exemplars from the training set to facilitate in-context learning for LLMs. 
We demonstrate the effectiveness of our approach on the long-term action anticipation (LTA) task, achieving state-of-the-art (SOTA) performance on Ego4D benchmarks.

\subsection{Task Description}

The LTA task involves predicting a chronological sequence of future actions given an untrimmed video $V$. The input video $V$ comprises multiple action segments with variable lengths, where each action segment corresponds to a single action (verb and noun pair). The segment boundaries for each action in the videos are provided as part of the input. Formally, after observing the video until time $t$ (\eg, $V_{:t}$), the LTA model sequentially predicts $Z$ future actions that the camera-wearer is likely to perform:
\begin{equation}
    \{\hat{n}_z, \hat{v}_z \}_{z=1}^Z 
\end{equation}
where $\hat{n}_z \in \mathcal{N}$ and $\hat{v}_z \in \mathcal{V}$, given the predefined noun and verb taxonomy $\mathcal{N}$ and $\mathcal{V}$. The input video $V_{:t}$ contains $P$ action segments. 

\subsection{Image Captioning Module}\label{subsec:image_captioning}

The past action labels do not capture all the visual information and context presented in the video (\eg., the location of the person or the background objects). Our idea is to use an image captioning model to capture the action context, providing more informative prompts related to the action performer for LLMs. 

\begin{figure}[h]
\centerline{\includegraphics[scale=.45]{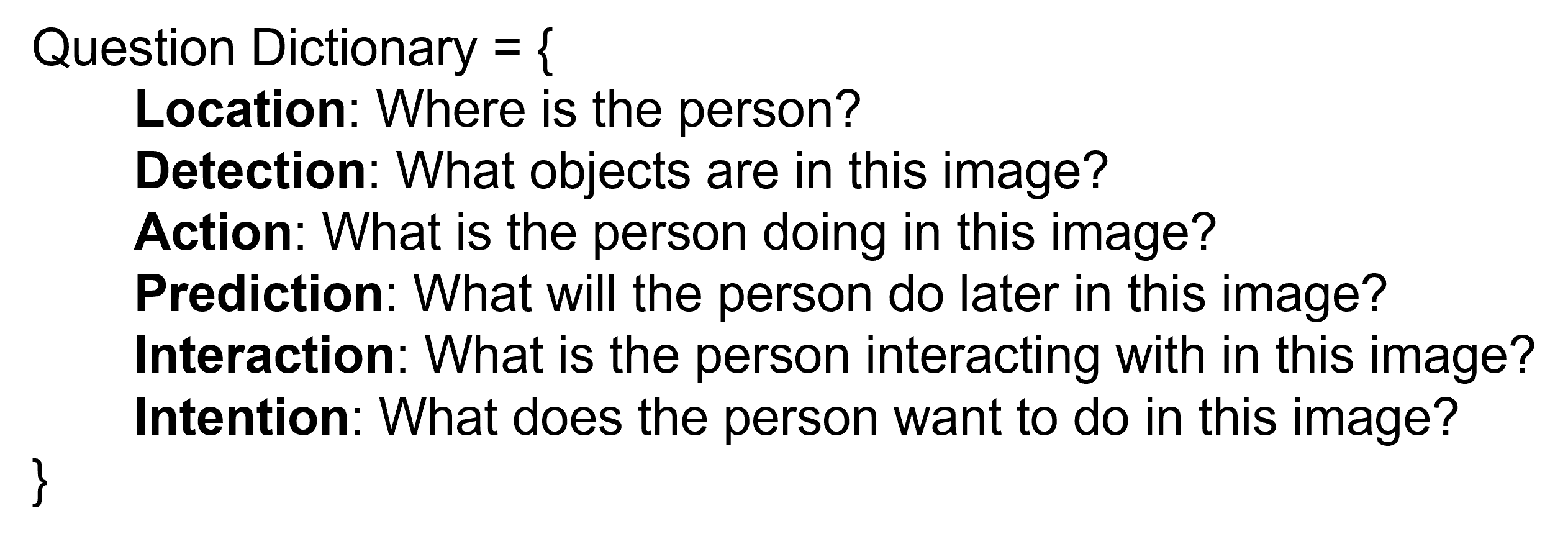}}
\vspace{-1em}
\caption{\textbf{List of questions} asking for information related to key concepts of egocentric image.}
\label{fig:question_dictionary}
\vspace{-2em}
\end{figure}

To enhance the descriptive power of captioning models, we formulate the image captioning task as a question-answering problem. \cref{fig:question_dictionary} displays a set of questions related to six key concepts that the models should aim to answer given an input image: location, detection, action, prediction, interaction, and intention. Given a question $Q$, the conditional text is formed as "Question: $<Q>$?, Answer: " where the answer part will be filled by the captioning model. After extensive experiments, we select intention-based question for conditional generation and VideoBLIP~\cite{VideoBLIP} as our image captioning model (see supplementary).

By posing these questions to the captioning models, we anticipate extracting useful descriptions beneficial for LLMs to predict future actions. In practice, the conditional caption generation yields consistent and accurate descriptions of the action context for later prediction. For image captioning, we use the middle frame of each clip segment to generate the captions, based on the assumption that the middle frame is highly likely to be the moment when the action is happening.

\subsection{Action Recognition Module}\label{subsec:action_recognition}

\begin{figure*}[h]
\centerline{\includegraphics[scale=.35]{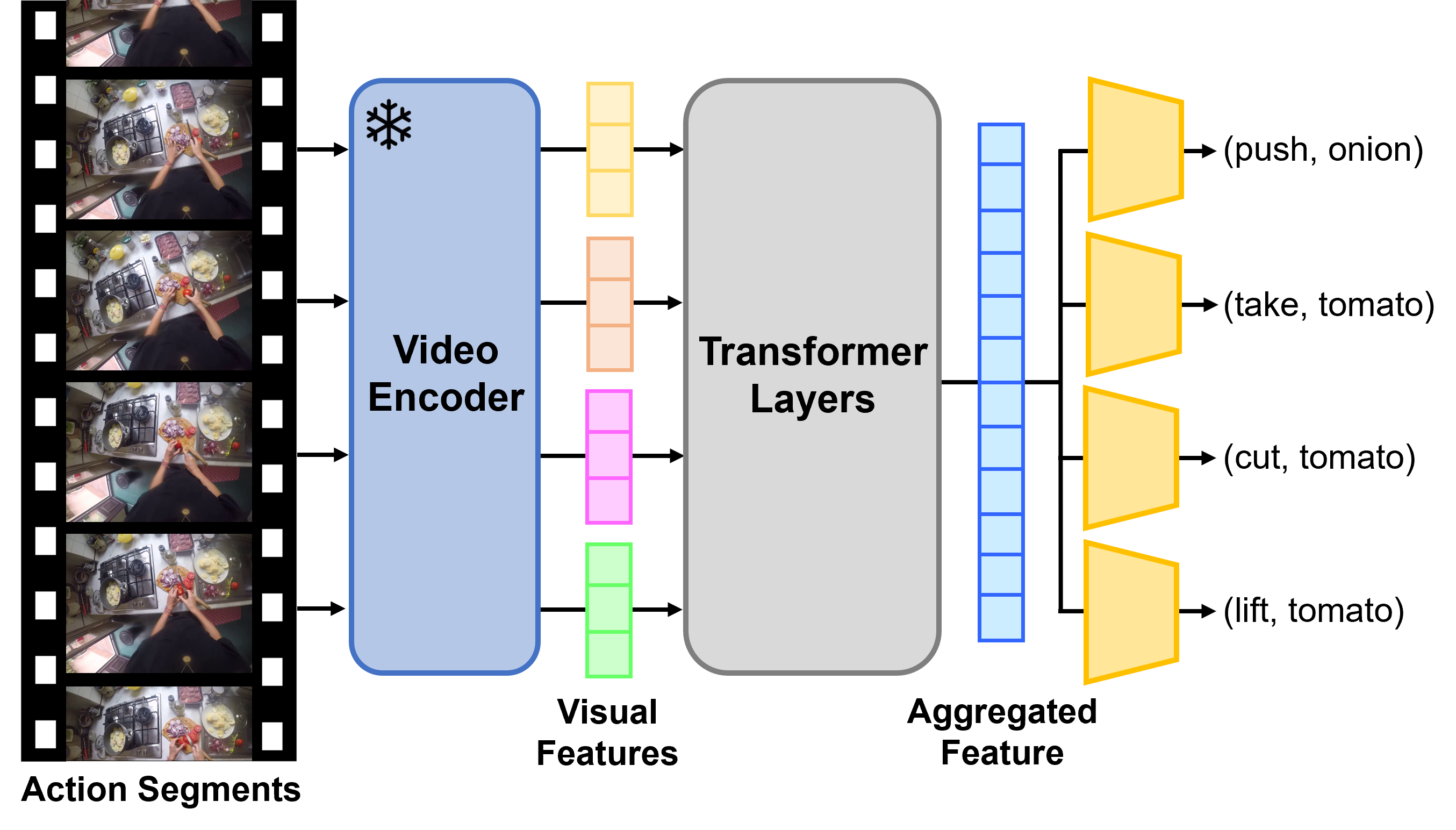}}
\caption{\textbf{Action recognition network}. A frozen video encoder generates visual features from $n=4$ continuous action segments, followed by transformer layers and classification heads. 
}
\label{fig:action_recognition}
\vspace{-0.7cm}
\end{figure*}

The accurate past action descriptions are crucial, as they provide essential contextual information about past events in a given video input. Our model recognizes actions from the observed video clips, taking inspiration from the baseline model used in the Ego4D paper~\cite{grauman2022ego4d}. 
Visual features are extracted from a video encoder given $n$ consecutive action segments ($n=4$ in \cref{fig:action_recognition}). 
We adopt the off-the-shelf video encoder of EgoVLP~\cite{lin2022egocentric}, a vision-language model pre-trained on Ego4D. Next, transformer layers are used to combine multiple visual features into an aggregated feature, followed by $n$ classification heads to recognize the verbs and nouns independently. The video encoder is frozen, and the rest of the model is trained with cross-entropy loss summed over $n$ classification heads. We also try to exploit a text encoder by formulating the action recognition task as matching a proper action label with an input video, which does not improve the accuracy (See supplementary).

We empirically find that recognizing actions in isolation falls short in effectiveness compared to recognizing $n$ actions within a continuous sequence.  This is because actions in neighboring segments are often highly correlated, reflecting the natural continuity embedded in human activities. Thus, we use a sliding input video window that includes $n=4$ segments with the window striding of one. The top-1 action is determined by identifying the most common action from the classifications across different windows. See supplementary for more details and ablation results.

\subsection{Action Anticipation Module}\label{subsec:large_language}

In this section, we discuss the details of action anticipation with LLMs, from how the prompt is formatted to the details of inference and post-processing of raw outputs. Llama 2 (7B)~\cite{touvron2023llama2}, an open-source LLM, is employed in our framework.

\begin{figure*}[h]
\centerline{\includegraphics[scale=.4]{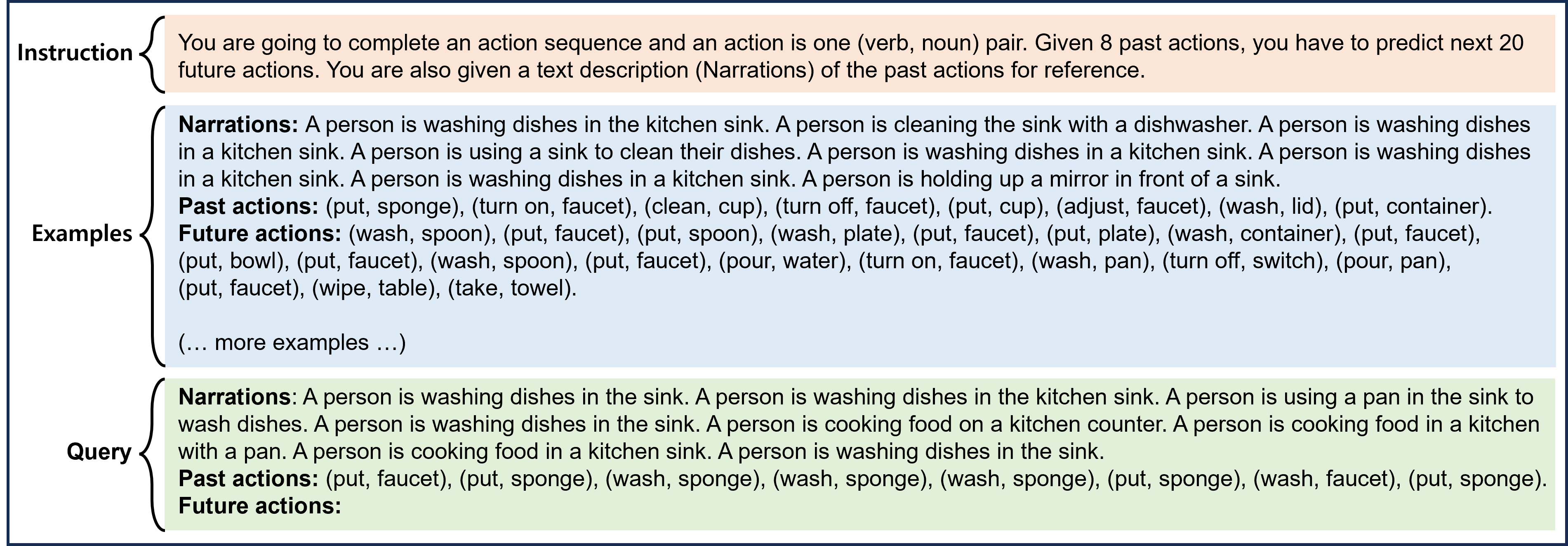}}
\caption{\textbf{Prompt structure for LLMs}. The prompt is composed of an \textcolor{pink}{instruction}, \textcolor{blue}{examples} retrieved from the training set, and the \textcolor{green}{query} related to the prediction video. Each video description consists of narrations, past actions, and future actions.}
\label{fig:prompt_template}
\vspace{-0.5cm}
\end{figure*}

\myparagraph{Prompt Design} Following common practice, we design the prompt with a few examples to enable in-context learning (ICL), maximizing references given to LLMs. \cref{fig:prompt_template} shows the template of our prompts, which consists of an instruction, a few examples, and a query. The instruction includes the task description and defines the input and output for LLMs. The examples for ICL, selected from the training set, include past and future actions (ground truth) and corresponding textual descriptions generated by the captioning model, which we call narrations. The query includes recognized past actions and generated narrations, and future actions are left empty for LLMs to complete.


\myparagraph{Example Selection} 
Incorporating some examples that demonstrate proper generation will greatly improve the performance of LLMs by providing additional context and references~\cite{naveed2023comprehensive, brown2020language}. Considering the diversity of egocentric videos, especially within the Ego4D dataset, randomly selected examples are unlikely to generalize well on various scenarios within the test set. 

Therefore, we utilize a maximal marginal relevance (MMR)~\cite{ye2022complementary, carbonell1998use} based exemplar selection strategy. The main idea behind MMR is to select exemplars that are relevant to the query while exhibiting sufficient variety to provide robust generalization across the test set. To be specific, we iteratively select a set of examples $p_i \in T$ from the training set $D$ that are semantically close to the query prompt $q$ but also diverse enough to provide additional information. 
\begin{equation}
\vspace{-0.1cm}
    p_j = \argmax_{p_j\in D/T} \left[\lambda S(q, p_j) - (1-\lambda) \max_{p_i \in T} S(p_i, p_j)\right],
\end{equation}
where $S$ measures the semantic similarity between exemplars and the parameter $\lambda$ strikes a balance between similarity and diversity. One exemplar includes past and future action sequences and corresponding captions in natural language. We use MPNet~\cite{song2020mpnet} to extract text embeddings and calculate cosine similarity to measure semantic similarity. We compare various exemplar selection strategies in \cref{sec:ablations}.

\myparagraph{Fine-tuning LLM} To facilitate LLMs to generate proper action sequence, we fine-tune the LLMs based on the training set of each egocentric dataset. As fine-tuning whole parameters of the LLMs can be computationally inefficient and may lead to catastrophic forgetting~\cite{kirkpatrick2017overcoming}, parameter-efficient fine-tuning is leveraged. Specifically, we utilize LoRA~\cite{hu2021lora} with 8-bit quantization using a batch size of 8 and a learning rate of $5 \times 10^{-4}$ for 10 epochs. The training set for fine-tuning consists of input prompts in the format of \cref{fig:prompt_template} with different exemplars and ground truth future actions as targets to be generated by the LLMs.

\myparagraph{Inference of LLM and Post-processing}
Following the evaluation protocol of Ego4d~\cite{grauman2022ego4d}, we generate multiple predictions for each test video. To reduce the variability in predictions, we prompt LLMs multiple times using the same prompt structure, as shown in \cref{fig:prompt_template}, but with different exemplars. After inference, we map the raw output of LLMs to the label space of predefined verbs and nouns. 
One should note that the output of LLMs may not always fall within the given taxonomy, despite all input prompts containing verbs and nouns from the predefined domain. For simplicity, we adopt a rule-based mapping (see details in supplementary).

\section{Experiments}\label{sec:experiment}

In this section, we evaluate \methodname on the long-term action anticipation (LTA) task, shedding light on the various design aspects. We conduct a comprehensive evaluation using the Ego4D dataset, which stands as the largest repository of egocentric videos. We report the final performance metrics on the Ego4D~\cite{grauman2022ego4d}, EPIC-KITCHEN-55~\cite{damen2020epic}, and EGTEA Gaze+~\cite{li2018eye} benchmarks.

\subsection{Datasets}

\myparagraph{Ego4D~\cite{grauman2022ego4d}} With 243 hours of egocentric video content encompassing 53 distinct scenarios, Ego4D offers a rich variety of data with a predefined annotation of 117 verbs and 521 nouns for the LTA task. Ego4D is available in two versions, v1 and v2, where Ego4D v1 is a subset of Ego4D v2 with fewer videos and a smaller verb and noun set. We use both versions of the dataset to enable extensive comparison with previous methods and adhere to the identical training, validation, and test splits as specified in the benchmark.

\myparagraph{EPIC-KITCHENS-55 (EK-55)~\cite{damen2020epic}} This dataset comprises 55 hours of egocentric videos centered around cooking scenarios, featuring 125 annotated verbs and 352 nouns. We follow the same train and test splits as EGO-TOPO~\cite{nagarajan2020ego}.

\myparagraph{EGTEA Gaze+ (EGTEA)~\cite{li2018eye}} With approximately 26 hours of egocentric cooking videos, EGTEA includes 19 verbs and 53 nouns. Similar to EK-55, we employ the same train and test splits as described in EGO-TOPO~\cite{nagarajan2020ego}.

\subsection{Comparison to the SOTA}~\label{subsec:comparison_to_SOTA}

\begin{table}[t]
\centering
\small
\setlength\tabcolsep{3pt}
\caption{\textbf{Comparison to the SOTA on the Ego4D test set.} \methodname outperforms all baselines on the Ego4D LTA benchmark in respect of edit distance (ED $\downarrow$).} 
\begin{tabular}{lcccc}
\toprule
Method                          & Version   & Verb ED       & Noun ED    & Action ED     \\ \midrule

Slowfast~\cite{grauman2022ego4d}& v1   & 0.7389          & 0.7800          & 0.9432           \\
VCLIP~\cite{das2022video+}       & v1   & 0.7389          & 0.7688          & 0.9412           \\
ICVAE~\cite{mascaro2023intention}& v1   & 0.7410          & 0.7396          & 0.9304           \\
HierVL~\cite{ashutosh2023hiervl}& v1   & 0.7239          & 0.7350          & 0.9276           \\
AntGPT~\cite{zhao2023antgpt}    & v1   & 0.6584 & 0.6546          & 0.8814          \\ 
\cdashline{1-5}
\methodname (ours)                     & v1   & \textbf{0.6559} & \textbf{0.6401} & \textbf{0.8613}  \\\midrule
Last action                     & v2   & 0.8329          & 0.7258          & 0.9438  \\
Repeat action                   & v2   & 0.7491          & 0.7073          & 0.9155  \\
Retrieve action                & v2   & 0.7417          & 0.7041          & 0.9127  \\
Slowfast~\cite{grauman2022ego4d}& v2   & 0.7169          & 0.7359          & 0.9253           \\
VideoLLM~\cite{chen2023videollm}& v2   & 0.721          & 0.725            & 0.921            \\
AntGPT~\cite{zhao2023antgpt}    & v2   & 0.6503          & 0.6498          & 0.8770          \\ 
\cdashline{1-5}
\methodname (ours)                     & v2   & \textbf{0.6471} & \textbf{0.6117} & \textbf{0.8503} \\\bottomrule
\end{tabular}
\label{tab:ego4d_final_results}
\end{table}

\vspace{-0.5cm}
\myparagraph{Baselines}
We include three baselines:  
\begin{itemize}
    \item \textbf{Last action} outputs the last action of past action sequence as $Z$ future actions.
    \item \textbf{Repeat action} repeats the given past $P$ actions to fill $Z$ future actions.
    \item \textbf{Retrieve action} retrieves the most semantically similar example in the training set (as in the example selection step) and uses their future actions as the prediction. 
\end{itemize}  
We also compare to three previous SOTA methods:
\begin{itemize}
    \item \textbf{Slowfast}~\cite{grauman2022ego4d} is the baseline LTA model given by Ego4d.
    \item \textbf{VideoLLM}~\cite{chen2023videollm} converts multimodal data into a unified token sequence and utilizes fine-tuned LLMs 
    \item \textbf{AntGPT}~\cite{zhao2023antgpt} leveraged action recognition model and fine-tuned LLMs to predict future actions.
\end{itemize}

\begin{figure}[H]
\centerline{\includegraphics[scale=.33]{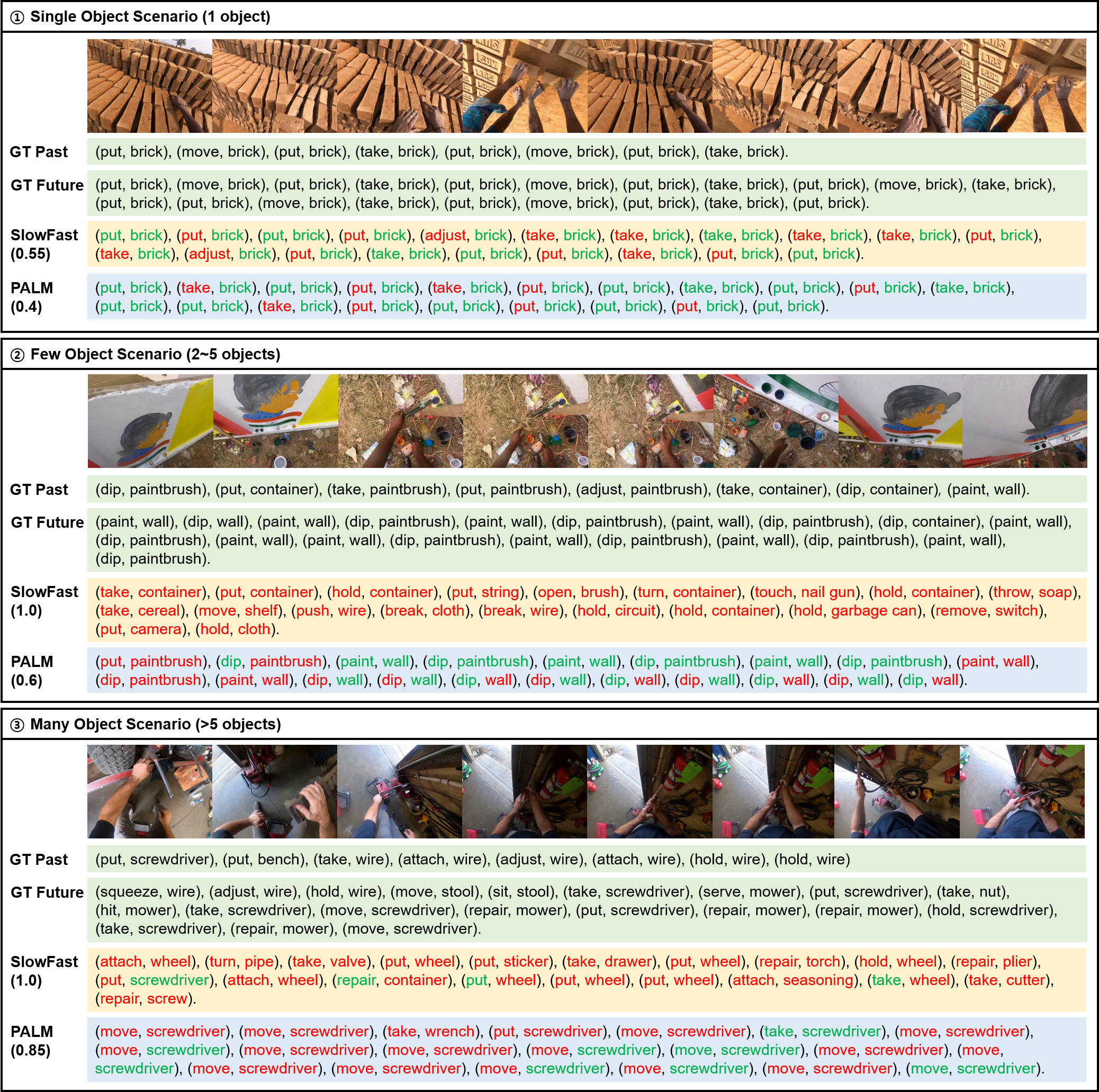}}
\caption{\textbf{Qualitative results of predictions on Ego4D.} We show the comparison between the predictions from the Ego4D baseline using \textbf{SlowFast}~\cite{grauman2022ego4d} and the predictions from our model \textbf{\methodname} respectively. For each of the three examples shown here, we include the middle frames of the previous eight action segments with the corresponding ground truth past actions \textbf{GT Past} and future actions \textbf{GT Future} alongside the predicted future actions. The \textcolor{green}{green} words represent correct predictions while the \textcolor{red}{red} words denote incorrect predictions. The edit distance between the predicted actions and ground truth actions is shown in parenthesis. }
\vspace{-0.5cm}
\label{fig:qualitative_results_prediction}
\end{figure}

\myparagraph{Evaluation}
We conform to the evaluation procedures set forth by the Ego4D~\cite{grauman2022ego4d} and calculate the edit distance (ED) using the Damerau-Levenshtein distance metric~\cite{damerau1964technique, levenshtein1966binary}. ED is computed separately for verbs, nouns, and the entire action sequences. Given $P=8$ action segments of the input video, we report the minimum edit distance among $K=5$ predicted sequences, each with a length of $Z=20$ actions. 

\myparagraph{Results}
\cref{tab:ego4d_final_results} shows that \methodname improves existing SOTA methods and achieves the best performance on the test set of the Ego4D v1 and v2 benchmark. In general, LLM-based approaches (VideoLLM, AntGPT, and \methodname) predict more accurate future actions compared to other approaches based on representation learning. Notably, repeat and retrieve action baselines in Ego4D v2 perform better than Slowfast with respect to noun and action ED. These observations clearly indicate the advantages of leveraging foundation models and the limitations of feature learning methods on the LTA task. Moreover, \methodname outperforms other LLM-based methods, which demonstrates the effectiveness of our approach to provide ample visual context to LLMs, equipped with action recognition and image captioning models.

In \cref{fig:qualitative_results_prediction}, we perform a qualitative comparison between the Ego4D baseline using SlowFast~\cite{grauman2022ego4d} and \methodname. 
The first example exemplifies a scenario where the person mainly interacts with a single object, in this case, a brick. Both methods can accurately predict the future noun ``brick''. 
In the second scenario, the person is drawing a painting on a wall, interacting with multiple objects such as a paintbrush, a container, and the wall. \methodname predicts the repeated sequence of actions, unlike SlowFast which incorrectly anticipates unrelated actions such as putting a string, touching a nail gun, or throwing a soap, none of which are present in the input video. 
This also demonstrates the importance of previous actions in the anticipation of future actions. 
The final scenario portrays the person adjusting a wire in a repair shop, a task involving numerous items.
While SlowFast and \methodname recognize central objects such as a screwdriver, they both fail to predict a coherent sequence of repair actions. 
We hypothesize that a language model with expert knowledge can produce more meaningful anticipations, particularly in the context of repairing mechanics, where it is crucial to execute actions in a precise order. 
More qualitative results are shown in supplementary.

\subsection{Ablations}
\label{sec:ablations}
We perform various ablation studies regarding action recognition, example selection methods, LLM models, and prompt contents. We evaluate on the validation set of Ego4D v2.

\begin{table*}[t]
\centering
\caption{ \textbf{Ablation on action recognition module.} We compare different pre-trained video encoders by reporting the Top-1 accuracy (Acc. $\uparrow$) and edit distance (ED $\downarrow$) for verbs, nouns, and actions. For comparative purposes, we also include the ED when ground truth actions (GT action) are used in the prompts. The best Acc. and ED are highlighted in bold except those from GT action. 
}
\begin{tabular}{lcccccc}
\toprule
Method      & Verb Acc.  & Noun Acc. & Action Acc.   & Verb ED          & Noun ED          & Action ED      \\ \midrule
MViT~\cite{fan2021multiscale}& 19.87          & 2.55           & 0.51           & 0.7291          & 0.8597          & 0.9630          \\
SlowFast~\cite{feichtenhofer2019slowfast}& 19.42          & 14.65          & 3.12           & 0.7252          & 0.7472          & 0.9278          \\
StillFast~\cite{ragusa2023stillfast}& 19.12          & 19.41          & 4.06           & 0.7356          & 0.7369          & 0.9241          \\
EgoVLPv2~\cite{pramanick2023egovlpv2}& 33.84          & 40.63          & 16.31          & 0.7182          & 0.6651          & 0.8937          \\
EgoVLP~\cite{lin2022egocentric}& \textbf{40.32} & \textbf{45.53} & \textbf{20.63} & \textbf{0.7111} & \textbf{0.6465} & \textbf{0.8819} \\
\cdashline{1-7}
GT action               & -              & -              & -              & 0.6663          & 0.5887          & 0.8319         \\  \bottomrule
\end{tabular}
\label{tab:action_recognition}
\vspace{-0.4cm}

\end{table*}

\myparagraph{Action Recognition}
The action recognition module is an essential part of our framework \methodname as we utilize LLMs for predicting future actions and the accuracy of these predictions is highly influenced by the sequence of previous actions. 
We evaluate various video encoders trained on egocentric videos, including MViT~\cite{fan2021multiscale}, SlowFast~\cite{feichtenhofer2019slowfast}, StillFast~\cite{ragusa2023stillfast}, EgoVLP~\cite{lin2022egocentric}, and EgoVLPv2~\cite{pramanick2023egovlpv2}. It is important to note that we employ a frozen video encoder without any fine-tuning, as depicted in \cref{fig:action_recognition}.

\cref{tab:action_recognition} presents the Top-1 Accuracy (\%) and the edit distance of each video encoder. EgoVLP emerges as the top performer, achieving the best accuracy and edit distance among all encoders. EgoVLP utilizes a custom contrastive loss to align egocentric video frames with narrations, effectively establishing a connection between visual and textual embeddings. However, EgoVLPv2 utilizes task-specific loss for question-answering tasks built on the EgoVLP setting, which results in slightly less informative video features for action recognition. Unlike EgoVLP and EgoVLPv2, other video encoders are solely trained on video frames, making it challenging to draw robust connections between their video features and textual information (verbs and nouns).
Based on these results, we select EgoVLP as our video encoder.

It is expected that better accuracy in action recognition leads to a reduced edit distance, as accurate estimation of past actions is crucial for predicting future actions. Notably, the use of ground truth (GT) actions significantly outperforms all video encoders, which highlights the enhancement of action recognition as the foremost challenge within our framework. We believe that fine-tuning the video encoder on specific egocentric datasets would likely enhance recognition accuracy. 

\begin{table}[h]
\centering
\caption{\textbf{Ablation on example selection strategies} exhibits that MMR outperforms other selection methods.}
\begin{tabular}{lccc}
\toprule
Method       & Verb ED         & Noun ED        & Action ED       \\ \midrule
Random & 0.7362          & 0.6747          & 0.9002          \\ 
Similarity & 0.7204          & 0.6661          & 0.8923          \\
MMR
& \textbf{0.7179}   & \textbf{0.6647}          & \textbf{0.8908}          \\ \bottomrule
\end{tabular}
\label{tab:exemplar_selection}
\end{table} 

\myparagraph{Example Selection} We compare different example selection strategies for in-context learning (ICL), including random, similarity, and maximal marginal relevance (MMR)~\cite{ye2022complementary}. The random method uniformly selects examples from the training set, whereas the similarity method selects the most semantically similar example to the query. MMR considers both the similarity and diversity of examples, striving for a balance between them. \cref{tab:exemplar_selection} demonstrates that MMR outperforms the other selection methods. 

\begin{table}[h]
\centering
\setlength\tabcolsep{3pt}
\caption{\textbf{Ablation on LLMs} shows that larger LLMs perform better on the LTA task.}
\begin{tabular}{lccc}
\toprule
Method       & Verb ED         & Noun ED         & Action ED       \\ \midrule
GPT-Neo (125M) & 0.7657          & 0.6867          & 0.9119          \\ 
GPT-Neo (1.3B) & 0.7314          & 0.6759          & 0.8970          \\
GPT-Neo (2.7B) & \textbf{0.7178}   & \textbf{0.6747}          & \textbf{0.8943}          \\
Llama 2 (7B) & 0.7179          & 0.6647          & 0.8908          \\
Llama 2 (13B) & \textbf{0.6958}          & \textbf{0.6634}          & \textbf{0.8867}          \\ \bottomrule
\end{tabular}
\label{tab:llm_ablation}
\end{table} 

\myparagraph{LLM Ablation} 
We ablate the impact of LLMs by comparing different open-source models, primarily GPT-Neo~\cite{gpt-neo} and Llama 2~\cite{touvron2023llama2} families. \cref{tab:llm_ablation} indicates that larger language models are more capable of predicting future actions with their improved capacity for generalization across novel scenarios. This observation aligns with the performance of LLMs on general NLP benchmarks. Taking into account the computational resources at our disposal, we select Llama 2 with 7B parameters.

\begin{table}[h]
\centering
\caption{\textbf{Ablation on the content of the prompt.} We use abbreviations for each content in the input prompt: $\mathcal{A}$ - action sequence, $\mathcal{C}$ - captions}
\begin{tabular}{cccc}
\toprule
Prompt Content       & Verb ED         & Noun ED         & Action ED       \\ \midrule
$\mathcal{A}$   & 0.7203          & 0.6688          & 0.8947  \\
$\mathcal{C}$   & \textbf{0.7088}     & 0.6901          & 0.9064  \\
$\mathcal{A}+\mathcal{C}$   & 0.7179 & \textbf{0.6647}  & \textbf{0.8908} \\ \bottomrule
\end{tabular}
\label{tab:prompt_content_ablation}
\end{table}

\myparagraph{Prompt Contents} We examine different content included in the input prompt to validate the contribution of each module to final performance. \methodname utilizes recognized past actions and generated image captions to compose prompts. The action sequence describes the temporal flow of activity in (verb, noun) format, while captions narrate visual information relevant to human. In \cref{tab:prompt_content_ablation}, the action sequence with captions ($\mathcal{A}+\mathcal{C}$) achieves the lowest action ED, which thereby is used in our framework. Notably, caption-only prompt ($\mathcal{C}$) reaches the lowest verb ED, showing that narrations are capable of capturing dynamic motions of human activity, while action-only prompt ($\mathcal{A}$) effectively detects key objects interacted with human supported by relatively lower noun ED. This indicates that action sequence and narrations provide a beneficial complement, thereby supporting our decision to incorporate both in the prompts.


\begin{table}
\centering
\small
\setlength\tabcolsep{3pt}
\caption{\textbf{Comparison to the SOTA on EK-55 and EGTEA.} Performance is evaluated using the mAP metric ($\uparrow$).} 

\begin{tabular}{lcccccc}
\toprule
\multirow{2}{*}{Method} & \multicolumn{3}{c}{\textbf{EK-55}} & \multicolumn{3}{c}{\textbf{EGTEA}}                                    \\
                        & ALL     & FREQ   & RARE   & ALL  & FREQ   & RARE                      \\ \hline
I3D~\cite{carreira2017quo}& 32.7    & 53.3   & 23.0   & 72.1 & 79.3   & 53.3                      \\
ActionVLAD~\cite{girdhar2017actionvlad}& 29.8    & 53.5   & 18.6   & 73.3 & 79.0   & 58.6                      \\
Timeception~\cite{hussein2019timeception}& 35.6    & 55.9   & 26.1   & 74.1 & 79.7  & 59.7                      \\
VideoGraph ~\cite{hussein2019videograph}& 22.5    & 49.4   & 14.0   & 67.7 & 77.1  & 47.2                      \\
EGO-TOPO~\cite{nagarajan2020ego}& 38.0    & 56.9   & 29.2   & 73.5 & 80.7   & 54.7                      \\
Anticipatr~\cite{nawhal2022rethinking}& 39.1    & 58.1   & 29.1   & 76.8 & 83.3  & 55.1                      \\ 
AntGPT~\cite{zhao2023antgpt}& 40.1    & 58.8   & \textbf{31.9}   & 80.2 & 84.8 & 72.9 \\
\cdashline{1-7}
\methodname (ours)         &  \textbf{40.4}    &  \textbf{59.3}   & 30.3  &  \textbf{80.7}  &  \textbf{85.0}  & \textbf{73.5} \\ \bottomrule
\end{tabular}
\label{tab:EK_55_EGTEA_final_results}
\end{table}

\subsection{Evaluation on EK-55 and EGTEA}~\label{subsec:final_results_EK_EGTEA}

\myparagraph{Evaluation}
For EK-55~\cite{damen2020epic} and EGTEA~\cite{li2018eye}, we adhere to the experimental setup established by EGO-TOPO~\cite{nagarajan2020ego} and employ mean average precision (mAP) for multi-label classification as the evaluation metric. The initial $R\%$ ($R = [25\%, 50\%, 75\%]$) of each video serves as input, with the objective of predicting the set of actions occurring in the remaining $(100-R)\%$ of the video. Notably, actions here are confined to \emph{verb} predictions only without \emph{noun}, following previous conventions. We calculate the mAP on the validation set averaged over all $R$, examining various performance aspects, such as all target actions (ALL), frequently occurring actions (FREQ), and rarely occurring actions (RARE).

\myparagraph{Results}
As shown in \cref{tab:EK_55_EGTEA_final_results}, our method achieves SOTA results in the ALL and FREQ section while exhibiting comparable performance in the RARE section. The results demonstrate the effectiveness of \methodname by capturing additional visual information with image captioning models. This aligns with the ablation study regarding prompt contents in \cref{sec:ablations} where captions contribute better on verb predictions. We observe that verb in RARE section are seldom represented in the captions, which results in relatively lower mAP in RARE section. Thus, we believe fine-tuning image captioning model or employing better video captioning model will further improve the performance accordingly.


\section{Conclusion}\label{sec:conclusion}

We present \methodname, a framework that leverages vision-language and large language models for long-term action anticipation. 
\methodname achieves the best performance on the Ego4D benchmark, demonstrating the effectiveness of performing the LTA task in the language domain and the benefits of using foundation models. 
Our model also shows competitive results on two other benchmarks, EK-55 and EGTEA. 
%
%
%
%
As a result, we are optimistic that our work will spur further research in using commonsense knowledge extracted from LLMs for video understanding.
%


\clearpage
\appendix

\section{Overview}

We provide additional details of action recognition module (\cref{sup_sec:action_recognition}), image captioning module (\cref{sup_sec:image_captioning}), and action anticipation module (\cref{sup_sec:action_prediction}). We also show more qualitative examples (\cref{sup_sec:qualitative_examples}).

\section{Action Recognition Module}\label{sup_sec:action_recognition}

This section elaborates on the detailed implementation of the action recognition module.

\subsection{Implementation Details} 

The effectiveness of action recognition is paramount in influencing the final action anticipation results within \methodname. 
We formulate the task as a classification task and aim to find the most suitable verb and noun labels, as illustrated in the main paper. The video encoder generates visual embeddings, which are aggregated and then provided to classification heads to predict verbs and nouns.
In our ablation study presented in the main paper, we select EgoVLP~\cite{lin2022egocentric} for the video encoder. 

EgoVLP is pre-trained on the vision-language dataset EgoClip, utilizing a customized contrastive loss. Enabling multimodal data training, EgoVLP employs video and text encoders, independently mapping input video and text to corresponding embeddings. A contrastive loss is then calculated based on the similarity matrix between video and text embeddings. The extensive pre-training dataset and well-crafted loss empower EgoVLP to perform effectively on downstream egocentric tasks.

Our action recognition network, as illustrated in the main paper, comprises a frozen video encoder from EgoVLP, utilizing the TimeSformer~\cite{bertasius2021space} architecture. The subsequent transformer layers and classification heads are directly adopted from the Ego4D baseline. Visual features from action segments are initially extracted through the frozen video encoder, and the remainder of the model is trained with cross-entropy loss summed over all classification heads.

Note that we have decided to not include InternVideoEgo4D~\cite{chen2022internvideoego} in our comparison as InterVideoEgo4D is very similar to the StillFast~\cite{ragusa2023stillfast} model. Both models demonstrate comparable performance in action forecasting within the Ego4D benchmark, making the inclusion of InternVideoEgo4D redundant for our purpose. 

StillFast was originally designed for the short-term object interaction anticipation task (STA), while our method is tailored for the long-term action anticipation task (LTA). Specifically, StillFast processes high-resolution still images and low-resolution videos simultaneously to extract video features efficiently. These features are then passed through the prediction head (faster R-CNN) to generate outputs for the STA task. For comparison with other video encoders, we leverage the video features from the pre-trained StillFast model, feeding them sequentially through transformer layers and classification heads (see Fig. 4 of the main paper).

Given that the Ego4D LTA and STA datasets largely overlap in video content with different configurations, we posit that the video encoder of Stillfast pre-trained on the STA video dataset can produce meaningful representations for the LTA task as well. Moreover, the architectural similarity between Stillfast and Slowfast, both containing a two-branch model for video input processing, led us to consider Stillfast as a suitable baseline for our task compared to Slowfast. 

\subsection{How to Determine Top-1 Action Label}

\begin{figure*}[h]
\centerline{\includegraphics[scale=.55]{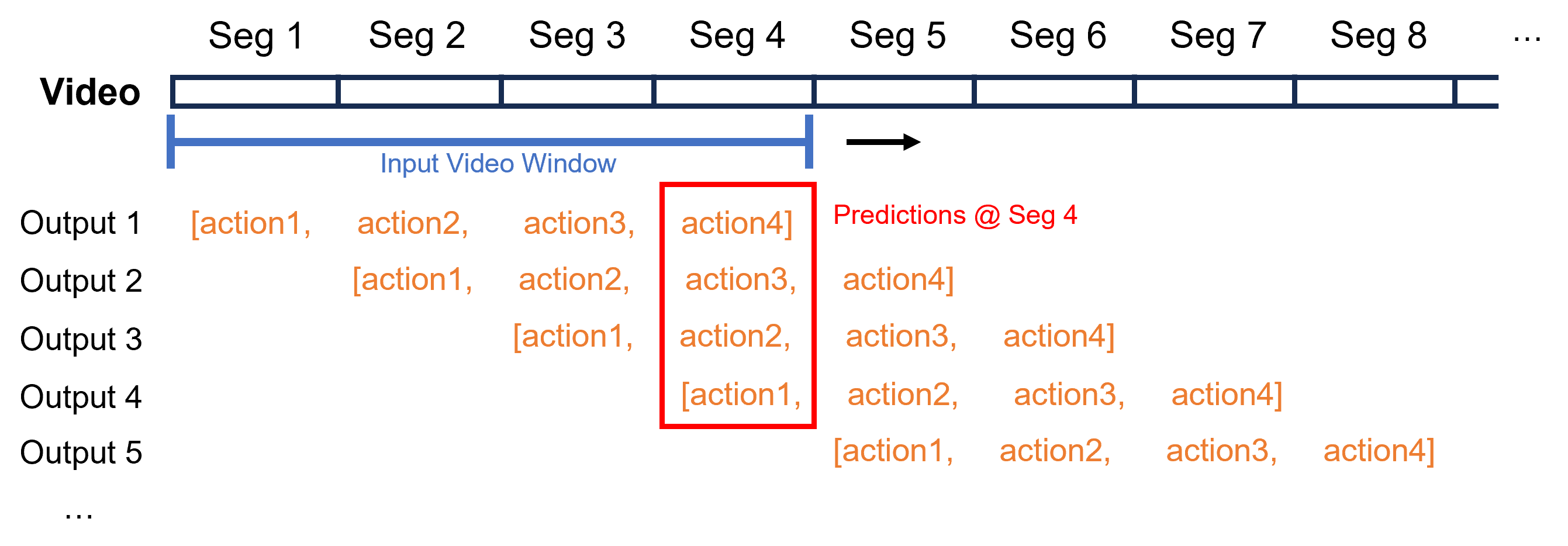}}
\caption{\textbf{A conceptual figure to describe how top-1 action is decided} among different outputs with input video windows. Input video consists of multiple action segments and our action recognition model classifies four actions with a sliding input video window. At Segment 4, all actions overlapped from Output 1 to Output 4 are aggregated and the most common verb and noun are decided as Top-1 action.}
\label{fig:action_top_1}
\end{figure*}

Our action recognition model accepts four consecutive action segments and generates four corresponding actions. 
This decision is made as adjacent segments are highly correlated, providing better insight into the sequence of visual dynamics. By this design choice, we can generate multiple outputs for a single segment with different video windows, as shown in \cref{fig:action_top_1}. Each output of the action recognition model consists of $K=5$ predictions generated by sampling based on the probability distribution of the classification head. Thus, the total number of actions at Seg 4 is $K*4 = 20$ and the most frequently appearing verb-noun pair is determined as the Top-1 action. We empirically find that our design results in better prediction results compared to one-to-one action recognition with deterministic sampling (\eg, choosing the verb and noun with the highest probability as the Top-1 action). 

\subsection{Ablation on the Number of Action Segments}

\begin{table}
\centering
\caption{\textbf{Ablation of the number of input segments for the action recognition model.} For one input segment, deterministic sampling is used while outputs are sampled $K=5$ times and aggregated for the number of input segments greater than one. Top-1 accuracy (Acc. $\uparrow$) is reported on the validation set of Ego4D v2.}
\begin{tabular}{cccc}
\toprule
\# of segment       & Verb Acc.         & Noun Acc.         & Action Acc.       \\ \midrule
1  & 38.37          & 40.63          & 17.02            \\ 
2  & 39.96          & 42.74          & 19.57           \\ 
4  & \textbf{40.32} & \textbf{45.53} & \textbf{20.63}   \\
8  & 38.67          & 41.31          & 17.57           \\ \bottomrule
\end{tabular}
\label{tab:action_recognition_input_clip_num}
\end{table}

\cref{tab:action_recognition_input_clip_num} demonstrates that the accuracy of action recognition depends on different numbers of input action segments. If the number of input segments is one, one segment generates one action output independently, similar to a typical classification problem. If the number of input segments is greater than one, the top-1 action is selected as depicted in \cref{fig:action_top_1} with sampling. It is evident that four consecutive action segments are optimal for capturing the temporal context of the video, resulting in the highest prediction accuracy. Our action recognition can be regarded as an ensemble method with different windows while embracing multiple actions with high probability through sampling.

\subsection{Video-Text Retrieval for Action Recognition}

\begin{figure}[h]
\centerline{\includegraphics[scale=.6]{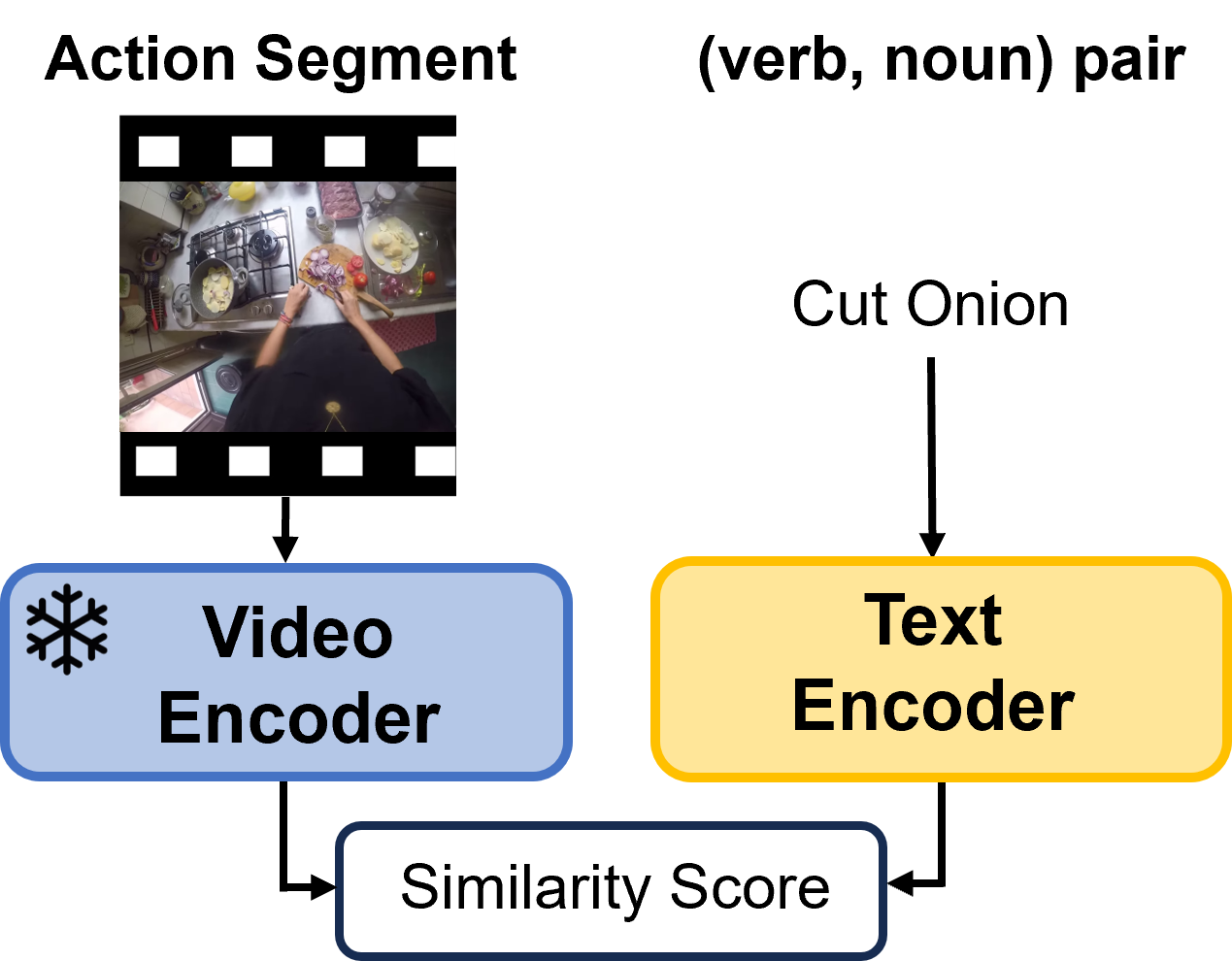}}
\caption{\textbf{Video-text retrieval for action recognition.} A frozen video encoder generates a visual feature from an action segment. The similarity score then is calculated by using the visual feature and textual feature produced by a text encoder, which is fine-tuned for the task.}
\label{fig:action_recognition2}
\vspace{-1em}
\end{figure}

When applying EgoVLP to this task, two design choices are available. First, we can utilize the video encoder of EgoVLP to solve action classification task as explained in the main paper. Second, the action recognition task can be formulated as matching a proper verb and noun pair (\eg, "wash dish") from a predefined dictionary to an input action segment as a video-text retrieval task. This can be implemented by using the text representation of verb and noun combinations as shown in \cref{fig:action_recognition2}. As the number of verb and noun combinations is limited, a similarity score is calculated between video embedding and text embeddings, generated by various combinations of verbs and nouns. The pair with the highest score is then selected. 

\begin{table}
\centering
\caption{\textbf{Comparison between (1) video-text retrieval-based and (2) classification-based (ours) approach for action recognition.} The retrieval-based approach (1) utilizes both the video and text encoders of EgoVLP while our approach (2) employs the video encoder, followed by transformer layers and classification heads. Top-1 accuracy (Acc. $\uparrow$) is reported on the validation set of Ego4D v2.}
\begin{tabular}{cccc}
\toprule
Method       & Verb Acc.         & Noun Acc.         & Action Acc.       \\ \midrule
(1)  & 32.87          & 38.26          & 15.18            \\ 
(2)  & \textbf{40.32} & \textbf{45.53} & \textbf{20.63}   \\ \bottomrule
\end{tabular}
\label{tab:egovlp_action_recognition_design_comparison}
\end{table}

We directly compare this retrieval-based approach with \methodname in \cref{tab:egovlp_action_recognition_design_comparison}. For setting (1), we fix the video encoder and finetune the text encoder on a predefined verb and noun set. To be specific, DistillBERT~\cite{sanh2019distilbert} is trained with the EgoNCE loss, a contrastive loss specially crafted by EgoVLP~\cite{lin2022egocentric}. In this way, we can avoid the expensive training of the video encoder. In \methodname (2), we utilize cross-entropy loss for each head to fine-tune the transformer layers and classification heads. 

\cref{tab:egovlp_action_recognition_design_comparison} demonstrates that the classification-based approach is more effective in terms of Top-1 accuracy than the video-text retrieval-based approach. 
This may be because the verb and noun pair (e.g., ``cut tomato") does not create a proper text description as EgoVLP is pre-trained with longer full-sentence narration (e.g., ``A person cuts a tomato to make a pasta in the kitchen."). Another interpretation is that only fine-tuning the text encoder while fixing the video encoder is not effective for multimodal training, as the parameters of both encoders should be updated properly according to contrastive loss.

\section{Image Captioning Module}~\label{sup_sec:image_captioning}
This section includes the ablation studies of image captioning models and captions conditioned on different questions.
\subsection{Ablation on Captioning Models}

\begin{table}[h]
\centering
\caption{\textbf{Comparison with different image captioning models} in terms of edit distance (ED $\downarrow$) on the validation set of Ego4D v2.}
\begin{tabular}{lccc}
\toprule
Method       & Verb ED         & Noun ED         & Action ED       \\ \midrule
BLIP~\cite{li2022blip}& 0.7201          & 0.6657          & 0.8927          \\
ViT-GPT-2~\cite{connectvit}& 0.7176          & 0.6680          & 0.8917          \\
GIT~\cite{wang2022git}& 0.7170 & 0.6667          & 0.8914          \\
InstructBLIP~\cite{instructblip}& 0.7182          & 0.6665          & 0.8909          \\
BLIP-2~\cite{li2023blip}& 0.7179          & 0.6647 & 0.8908 \\ 
VideoBLIP~\cite{VideoBLIP}& \textbf{0.7143}    & \textbf{0.6643} & \textbf{0.8907} \\ \bottomrule
\end{tabular}
\label{tab:captioning_comparison}
\end{table} 

The role of image captioning in providing additional context to the input prompt, beyond the past action sequence, is crucial for \methodname. Various image captioning models, including BLIP~\cite{li2022blip}, ViT-GPT-2~\cite{connectvit}, GIT~\cite{wang2022git}, InstructBLIP~\cite{instructblip}, BLIP-2~\cite{li2023blip}, and VideoBLIP~\cite{VideoBLIP}, are evaluated through ablation studies. All captioning models are utilized off-the-shelf without specific fine-tuning.

\cref{tab:captioning_comparison} illustrates that variations in captioning models have a marginal effect on the final edit distance, especially compared to the impact of the action recognition model. 
Notably, VideoBLIP, fine-tuned on the Ego4D benchmark, achieves the lowest action ED, directly extracting text information from the video, given a specific question. Considering the generalization ability and the conciseness of the generated captions, we select VideoBLIP as our captioning model. The captions generated by VideoBLIP accurately describe and effectively summarize visual information. 

\subsection{Question-Answering for Caption Generation}
\cref{tab:qa_caption} presents the edit distance measured with captions generated by different questions, including baseline captions generated with a prefix ("A person is"). Notably, intention, interaction, and prediction-related questions achieve lower edit distance compared to others. This aligns with the intuition that these questions require models to comprehend the purpose of human activity and the main objects of interest to users. Intention-based question is applied to generate captions in our framework as it achieves the lowest action edit distance.

\begin{table}[h]
\centering
\caption{\textbf{BLIP-2 captions generated with
different questions.} Edit distance (ED $\downarrow$) is calculated on the validation set of Ego4D v2.}
\begin{tabular}{lccc}
\toprule
Question     & Verb ED         & Noun ED        & Action ED       \\ \midrule
Baseline   & 0.7179          & 0.6647           & 0.8908 \\ \cdashline{1-4}
Location    & 0.7115          & 0.6644          & 0.8888          \\ 
Detection   & \textbf{0.7080} & 0.6645          & 0.8887          \\
Action      & 0.7111          & 0.6622          & 0.8885          \\
Prediction  & 0.7103          & 0.6626          & 0.8882          \\
Interaction & 0.7092          & 0.6642          & 0.8878          \\
Intention   & 0.7090          & \textbf{0.6613} & \textbf{0.8877} \\ \bottomrule
\end{tabular}
\label{tab:qa_caption}
\end{table} 

\vspace{-0.5cm}

\section{Action Anticipation Module}~\label{sup_sec:action_prediction}

In this section, we provide a detailed explanation of post-processing and present ablation studies on the number of exemplars.

\subsection{Post-processing of the Output of LLMs}
From the generated texts, we extract verbs and nouns within the predefined label space. Although our input prompts consistently contain verbs and nouns within the predefined domain, the output is not guaranteed to do so. To facilitate the transition from raw output to verb and noun labels, we employ a rule-based mapping.

The output string is initially parsed up to the first period ".", LLMs sometimes generate additional fictional examples assuming the current one is complete. Subsequently, following the input format, we further separate the output by commas and parentheses, attempting to match verbs and nouns from the label space with the output words. If multiple verbs or nouns are found, the longer word is selected. For example, if "turn on" and "turn" are both found, "turn on" becomes the final label. Another scenario is choosing "tape measure" over "tape." 

Additionally, when LLMs fail to generate proper words, indicated as "\_\_\_" in the output, we use the previous verb or noun to fill in the empty prediction, assuming it is likely the correct action. Moreover, we utilize a padding strategy for instances when the generated action sequence is shorter than the required $Z$ actions. We observe that different padding methods only have a marginal effect on the overall performance, so we decide to repeat the last action until we have a total of $Z$ predicted actions.

\subsection{Ablation on the Number of Examplars}

\begin{table}[h]
\centering
\caption{\textbf{Comparison of the number of exemplars} included in the prompt. Edit distance (ED $\downarrow$) is reported on the validation set of Ego4D v2.
}
\begin{tabular}{cccc}
\toprule
\# of input ex.       & Verb ED         & Noun ED         & Action ED       \\ \midrule
1   & 0.7179          & 0.6647          & 0.8908 \\ 
2   & 0.7078          & 0.6589          & 0.8861 \\ 
4   & \textbf{0.7050}   & \textbf{0.6588}  & \textbf{0.8859} \\ \bottomrule
\end{tabular}
\label{tab:ICL_input_ex_num}
\end{table}

According to Brown \etal~\cite{brown2020language}, the number of exemplars included in the prompt affects the quality of the generated text. Therefore, we measure edit distance using different numbers of exemplars selected by MMR~\cite{ye2022complementary}. Llama 2 (7B) is used for experiments with the prompt format described in the main paper.

\cref{tab:ICL_input_ex_num} demonstrates that increasing the number of examples for ICL actually improves the edit distance of predicted actions. There is a larger improvement from one to two exemplars than from two to four exemplars. Due to the limited input token of Llama 2 (around 4k), we cannot conduct experiments with more than four exemplars. The final results of the Ego4D test set in the main paper are generated with four exemplars.

\section{Qualitative Examples}~\label{sup_sec:qualitative_examples}

\begin{figure*}[t]
\centerline{\includegraphics[scale=.34]{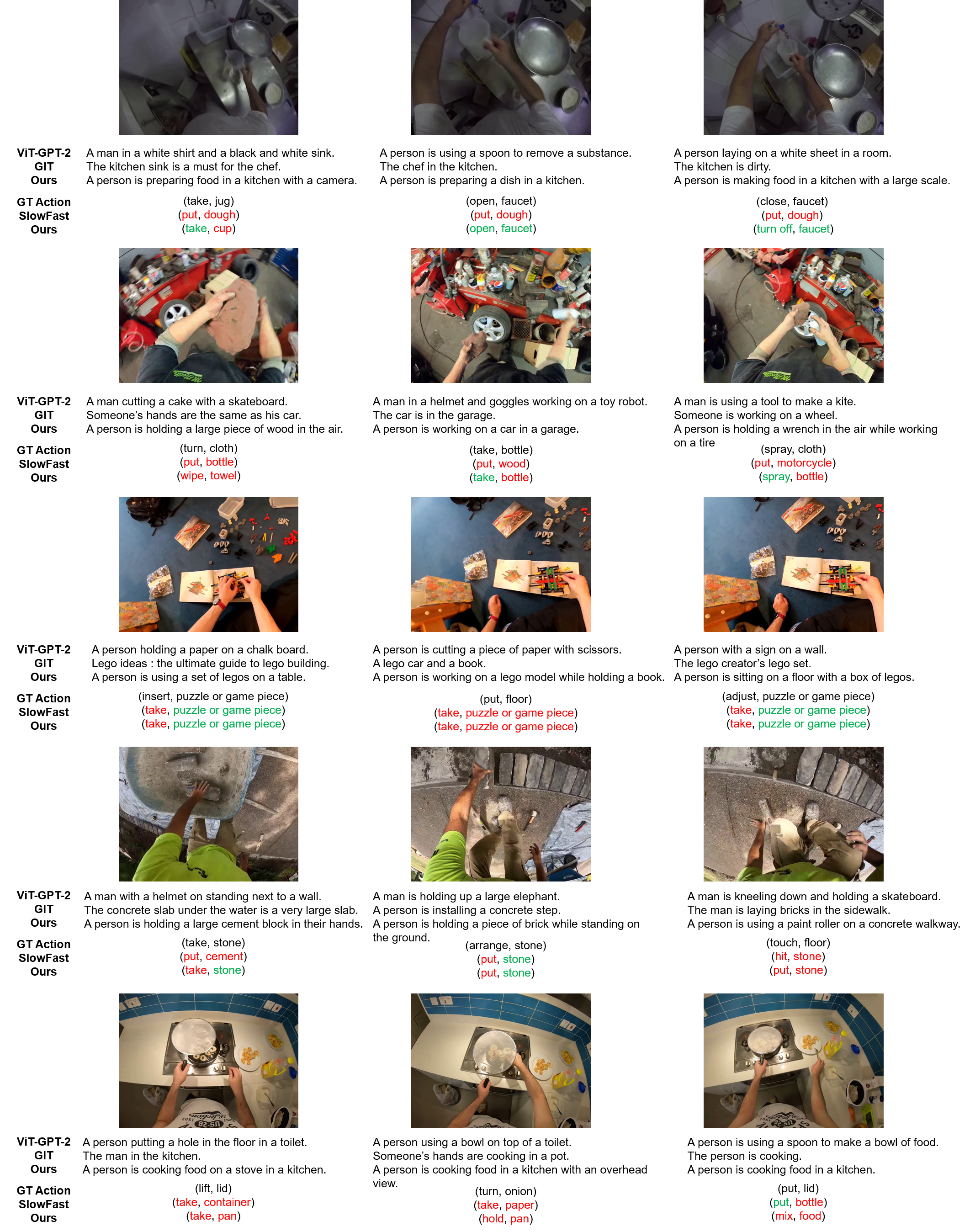}}
\caption{\textbf{Qualitative results on image captioning models and action recognition models}}
\label{fig:qualitative_results_action_caption}
\end{figure*}

\begin{figure*}[t]
\centerline{\includegraphics[scale=.34]{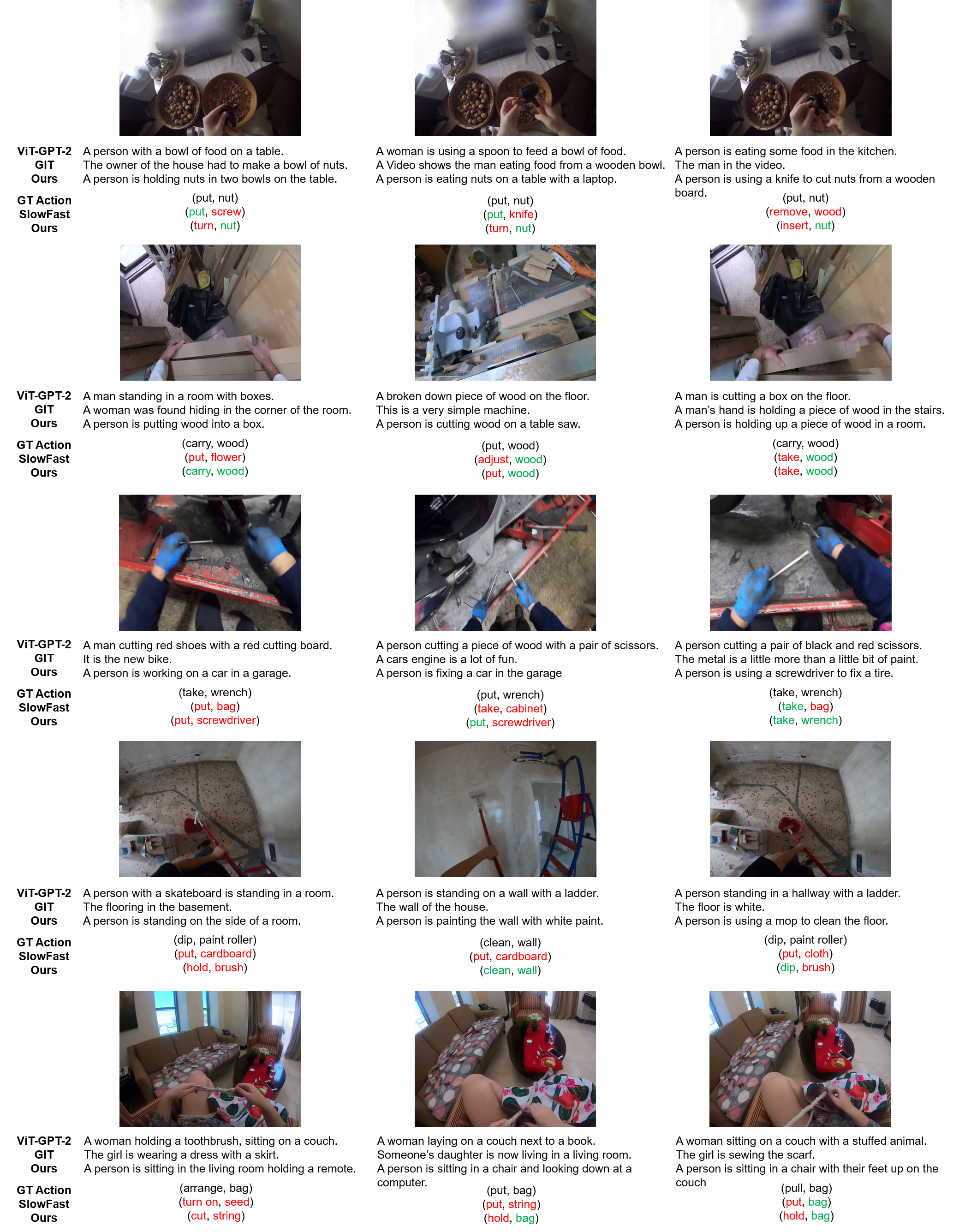}}
\caption{\textbf{Qualitative results on image captioning models and action recognition models.}}
\label{fig:qualitative_results_action_caption2}
\end{figure*}

\begin{figure*}[t]
\centerline{\includegraphics[scale=.34]{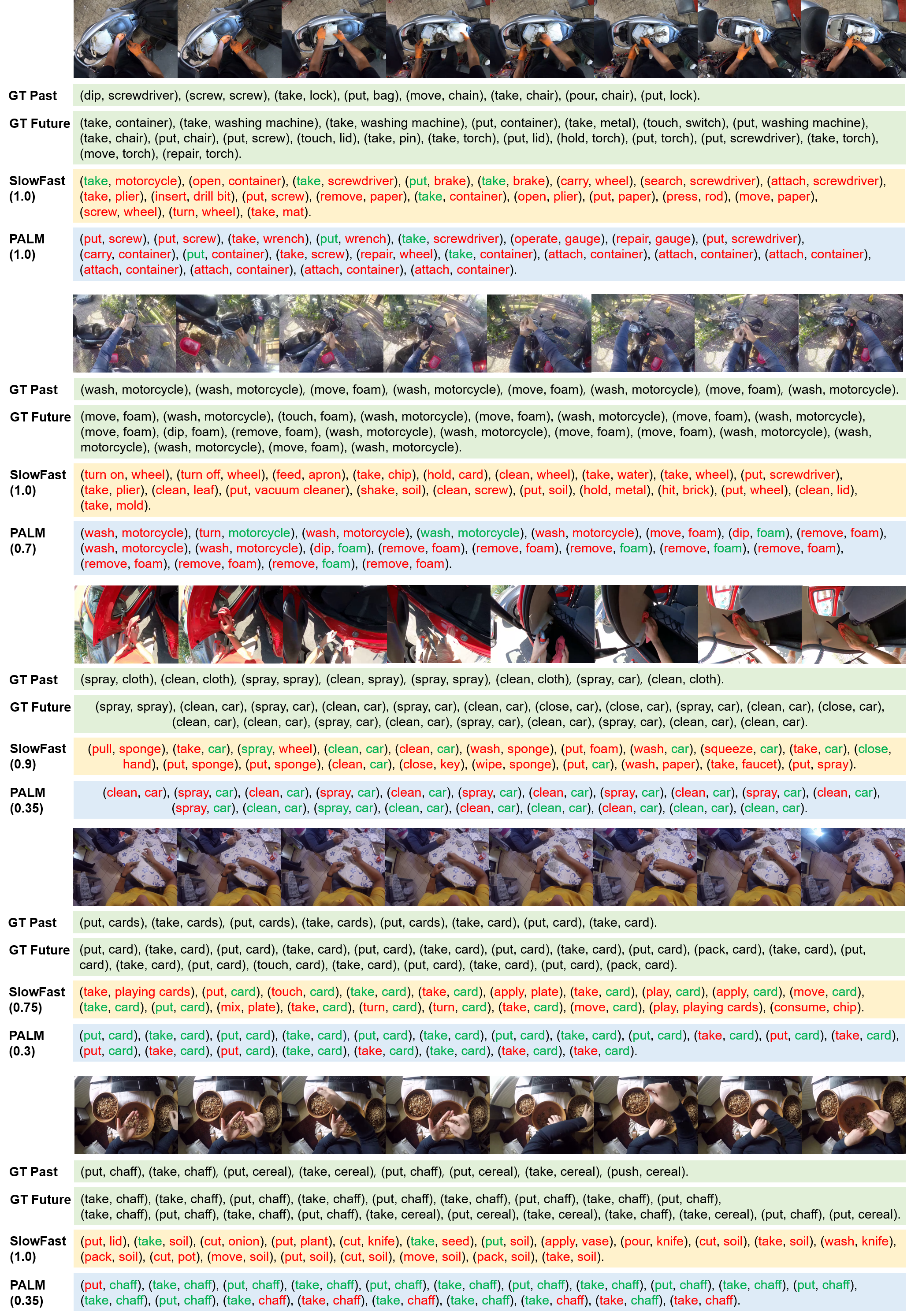}}
\caption{\textbf{Qualitative results of \methodname.}}
\label{fig:qualitative_results_prediction2}
\end{figure*}

\begin{figure*}[t]
\centerline{\includegraphics[scale=.34]{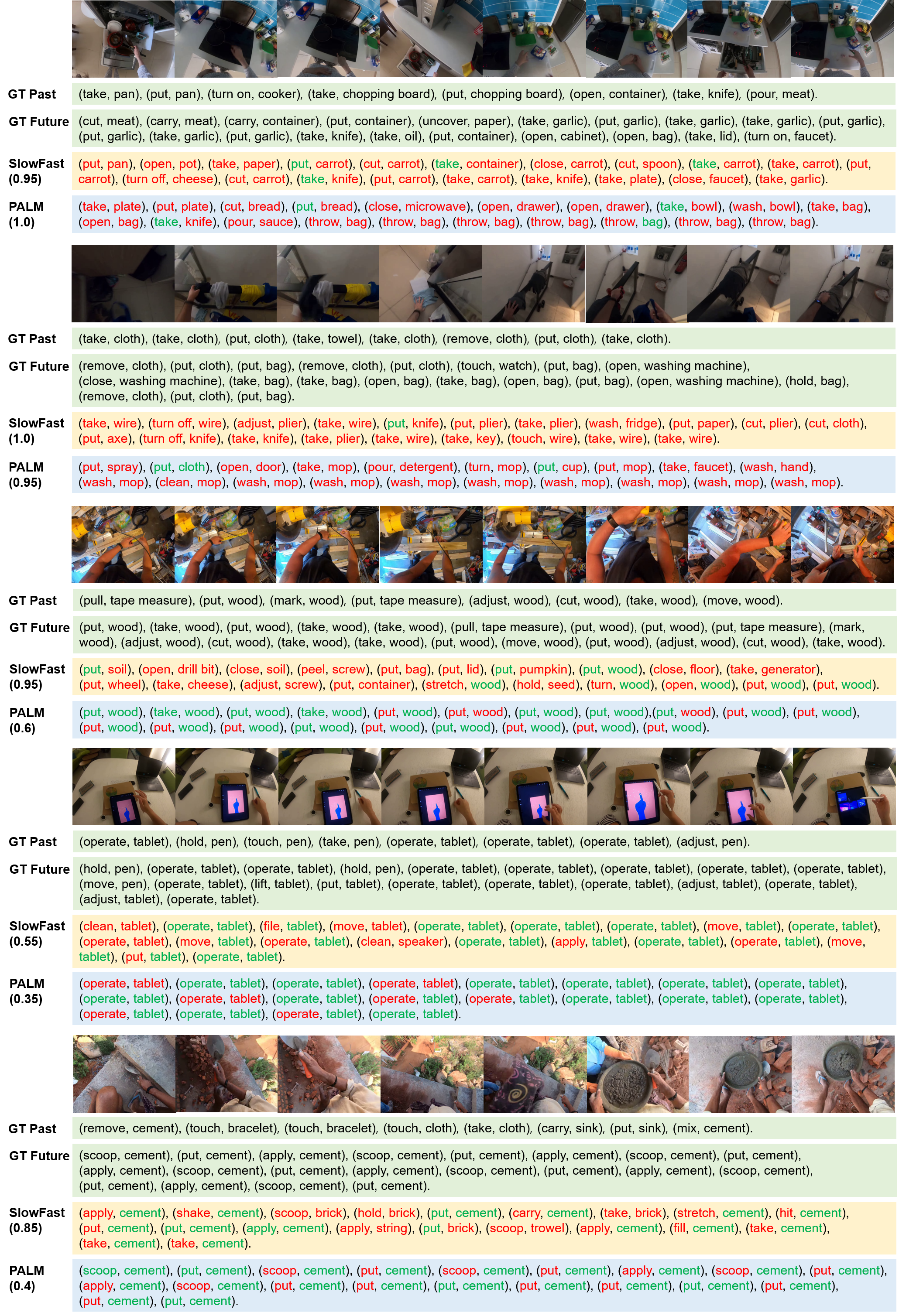}}
\caption{\textbf{Qualitative results of \methodname.}}
\label{fig:qualitative_results_prediction3}
\end{figure*}

In this section, we present various qualitative results. \cref{fig:qualitative_results_action_caption} and \cref{fig:qualitative_results_action_caption2} illustrate the qualitative results of action recognition and image captioning models. employing VideoBLIP and EgoVLP in our framework. Subsequently, \cref{fig:qualitative_results_prediction2} and \cref{fig:qualitative_results_prediction3} show more qualitative results of action predictions, comparing with the SlowFast~\cite{grauman2022ego4d} baseline.

\clearpage

%
%
\bibliographystyle{splncs04}
\bibliography{main}
\end{document}